%% file: acl_latex.tex
\documentclass[11pt]{article}

\usepackage[final]{acl}

\usepackage{times}
\usepackage{latexsym}
\usepackage{easy-todo}
\usepackage{hyperref}
\usepackage[T1]{fontenc}

\usepackage[utf8]{inputenc}

\usepackage{microtype}

\usepackage{inconsolata}

\usepackage{graphicx}
\usepackage{color}
\usepackage{booktabs}
\usepackage{multicol,multirow}

\usepackage[whole]{bxcjkjatype}

\usepackage{tcolorbox}

\tcbuselibrary{breakable}

\usepackage{fvextra}

\usepackage{makecell}
\usepackage{amsmath}

\usepackage{cgloss4e}
\usepackage{gb4e}
\noautomath

\title{Japanese Discourse Structure Analysis Based on Segmented
Discourse Representation Theory Using LLMs}

\author{
  \textbf{Satomu Tomisaki\textsuperscript{1,2}},
  \textbf{Yosuke Mikami\textsuperscript{1,2}},
  \textbf{Amane Watahiki\textsuperscript{1}},
  \textbf{Hitomi Yanaka\textsuperscript{1,2,3}}
\\
  \textsuperscript{1}The University of Tokyo,
  \textsuperscript{2}RIKEN,
  \textsuperscript{3}Tohoku University
\\
  \small{\texttt{\{tomisaki-no-mail, ymikami, hyanaka\}@is.u-tokyo.ac.jp, amanew@g.ecc.u-tokyo.ac.jp}}
}

\begin{document}
\maketitle
\begin{abstract}
Discourse parsing based on Segmented Discourse Representation Theory (SDRT) is crucial for capturing the structural dynamics of dialogues.
However, most prior research focuses on English, and its effectiveness for Japanese remains unclear owing to the lack of Japanese SDRT corpora.
To address this issue, we construct Japanese SDRT corpora by translating existing English SDRT corpora using an LLM.
Through a comparative evaluation of prompt-based and fine-tuned LLMs, we show that the models achieve comparable performance between English and Japanese, suggesting the cross-lingual applicability of SDRT-based discourse parsing.
In order to investigate why the English and Japanese parsers achieve similar performance, we conduct ablation analyses on discourse markers and contextual inputs. 
Our results reveal that most of the Japanese discourse markers we examined have little impact, and the models' predictions depend heavily on contextual discourse relations rather than on the content of the context clauses.
This structural reliance might explain the comparable performance despite the substantial typological differences between English and Japanese.

\end{abstract}
\section{Introduction}

Understanding a dialogue requires one to comprehend not only the meaning of individual discourse units but also how they are logically and pragmatically related.
Discourse parsing is the task of identifying such relations between individual discourse units.
It benefits numerous downstream tasks, such as dialogue summarization~\citep{meeting-summarization_gao-etal-2023-dialogue}, question answering~\citep{QAtask_he-etal-2021-multi}, and AI-generated text detection~\citep{AIDetection_kim-etal-2024-threads}.

To formalize these discourse relations, several theoretical frameworks have been developed.
Segmented Discourse Representation Theory (SDRT;~\citealp{SDRT}) is a prominent example.
Compared to other discourse analysis frameworks, SDRT is grounded in formal semantics and can handle multiparty conversations.

However, automating SDRT parsing remains computationally challenging.
Recent studies have explored simple prompt-based approaches~\citep{chan-etal-2024-exploring_prompt, fan-etal-2024-uncovering_promtmethod1}.
Zero- or few-shot prompting with models such as ChatGPT has not achieved satisfactory performance, leaving considerable room for improvement.
In contrast, models trained from scratch or fine-tuned on task-specific data have achieved high SDRT parsing performance in English~\citep{BERTLine,fan-etal-2025-improving,liu-etal-2025-enhancing,llamipa}.

Although SDRT is intended to be language-independent, existing SDRT resources and parsing studies predominantly focus on English. It therefore remains unclear how well existing SDRT parsing methods generalize to other languages. 
Languages differ in their lexical inventories and syntactic structures and may consequently use different cues to signal discourse relations. 
We hypothesize that parsers for different languages may rely on different information and achieve different levels of performance.
We take Japanese, which differs substantially from English in both lexicon and syntax, as a test case. 
For example, Japanese sentence-final particles may provide important cues for predicting discourse relations because these particles express the speaker's attitude toward information~\citep{BN04010577}. 
Differences in the use of connectives and sentence-final particles may therefore affect SDRT parsing performance.

However, to the best of our knowledge, no Japanese SDRT corpus is available.
This makes it difficult to train Japanese discourse relation parsers.
Without these parsers, we cannot study to what extent existing English SDRT parsing methods can be applied to Japanese, and what kind of information is crucial for Japanese discourse analysis.

In order to study the linguistic generalizability of SDRT, we first fill the resource gap.
We construct Japanese SDRT corpora by translating existing English SDRT corpora automatically. 
The translated corpora will be made publicly available for research purposes\footnote{Our corpora and experimental code will be available at \url{https://github.com/ynklab/Japanese-Discourse-Analysis}}.

Using these corpora, we examine how well SDRT-based parsing performs across languages.
We first evaluate LLM-based SDRT parsers on these translated corpora.
We consider two kinds of parsers: fine-tuned ones and prompt-based ones. 
We then compare the best-performing Japanese and English models.

Finally, we study the information that is crucial for identifying discourse relations.
We conduct three ablation analyses.
The first one is on two types of Japanese discourse markers: connectives and sentence-final particles.
The second one is on the previous discourse relations of the input.
The third one is on the content of contextual clauses.

Our results show that parsers perform comparably between English and Japanese.
The results of the first ablation show that, while most of the discourse markers we studied contribute little to the prediction of discourse relations,  \textit{demo} (でも; ``but'') has a substantial impact on identifying \textit{Contrast} relations.
The results of the second and third ablations show that model predictions primarily depend on the previous discourse relations, instead of the content.
The models' dependence on contextual discourse relations may explain the small performance gap between English and Japanese.
Because our Japanese corpora are translations of the English ones, these discourse relations are identical across the two languages, allowing the models to exploit the same information. 
As a result, they achieve comparable performance in both languages.

\section{Background}

\subsection{Discourse Analysis Framework}

Segmented Discourse Representation Theory (SDRT;~\citealp{SDRT}) is a theoretical framework for analyzing discourse structure.
In SDRT, a discourse is considered to be a sequence of elementary units.
There are two types of elementary units: Elementary Discourse Units (EDUs) and Elementary Event Units (EEUs).
Additionally, SDRT defines a compound unit, called a Complex Discourse Unit (CDU), which consists of multiple elementary units.
While an EDU refers to a clause in the text or conversation, an EEU refers to a salient nonlinguistic action or event in the surrounding context.
The entire discourse can then be represented as a directed acyclic graph (DAG), whose nodes are EDUs, EEUs, or CDUs, and the edges represent discourse relations among those units.
Figure~\ref{fig:MSDC} presents an example of such a discourse graph.

Compared with other frameworks, such as Rhetorical Structure Theory (RST;~\citealp{RST}), SDRT has two main advantages.
First, discourse relations in SDRT are defined more rigorously based on formal semantics.
Second, SDRT can be used to analyze multiparty conversations.
\input{figures/MSDC_figure}

\subsection{SDRT parsing with LLMs}
\label{SDRT parsing with LLMs}
Like many other areas of computational linguistics, discourse relation parsing is increasingly being tackled with LLMs.
In the context of SDRT parsing, approaches can be categorized into two groups: prompt-based methods and fine-tuning methods.
Existing prompt-based methods have not reached satisfactory performance; both zero- and few-shot prompting leave substantial room for improvement~\citep{chan-etal-2024-exploring_prompt, fan-etal-2024-uncovering_promtmethod1}.
In contrast, fine-tuning methods achieved higher parsing performance.
For example, BERTLine ~\citep{BERTLine} is a fine-tuned BERT.
It employs a multi-task architecture that simultaneously learns the attachments between EDUs and EEUs, as well as the discourse relations holding between them.
BERTLine outperformed previous neural models.
More recently, \citet{liu-etal-2025-enhancing} and \citet{fan-etal-2025-improving} have proposed methods with complex multi-module architectures, in which each module is built around a fine-tuned LLM.
Unlike the multi-module methods, Llamipa is a straightforward method that has a single module but shows competitive performance~\citep{llamipa}.
In this study, we select Llamipa as our primary fine-tuned parser because of its simplicity and efficiency.

Llamipa takes three inputs: (i) a target clause, (ii) its context, i.e., the preceding clauses within a fixed window, and (iii) the discourse relations among the context clauses.
A clause in the context is either an EDU or an EEU.
Llamipa uses a window size of 15, since discourse relations in SDRT corpora in general rarely span more than 15 clauses~\citep{llamipa}.
Llamipa then identifies which context clauses are related to the target clause and determines the discourse relation between each such clause and the target clause.

Llamipa incrementally predicts discourse relations.
Initially, Llamipa receives the first clause and therefore an empty set of discourse relations.
It then predicts the relation between the second clause and the first clause.
At each subsequent step, the previously processed target clause is added to the context, and the next clause becomes the new target.
Meanwhile, the discourse relations between the previous target clause and the clauses in the previous context are added to the discourse relation set.

There are two settings for using Llamipa, depending on which kind of relations we use to update the discourse relation set.
In the~\textbf{gold structure setting}, we use the gold relations.
In the~\textbf{predicted structure setting}, we use the relations predicted by Llamipa.
In the predicted structure setting, Llamipa iterates over the entire discourse and eventually constructs the entire discourse structure.

\subsection{SDRT Corpus}
\label{sec:SDRT_corpus}
In this paper, we use two English SDRT corpora: \textbf{Minecraft Structured Dialogue Corpus}~\citep[\textbf{MSDC};][]{MSDC} and \textbf{the Strategic Conversation}~(\textbf{STAC}) corpus~\citep{STAC}.

MSDC is the Minecraft Dialogue Corpus (MDC;~\citealp{MDC}) annotated with SDRT-based discourse relations\footnote{
The description of the relations is provided in Appendix~\ref{App:semantics}.
}.
MDC contains conversations between two players, a ``Builder'' and an ``Architect'', as they collaborate on tasks in the game \textit{Minecraft}.
Specifically, the Architect provides instructions to the Builder.
The Builder carries out these instructions and constructs specified structures.
Chat messages exchanged between the two players are treated as EDUs, and the descriptions of the Builder's actual behaviors (e.g., picking up or placing blocks) in the game are treated as EEUs.
Figure \ref{fig:MSDC} shows the discourse graph of an extract from MSDC. 

STAC is a corpus of chat logs and in-game events from the online version of the board game \textit{The Settlers of Catan}, all of which are manually annotated according to SDRT. 
In \textit{The Settlers of Catan}, players aim to accumulate victory points by acquiring and trading resources to construct roads, settlements, and cities. 
In this context, players use the chat interface to negotiate resource exchanges.
In STAC, chat messages are treated as EDUs and game information (e.g., dice rolls, card plays) are treated as EEUs.

\section{Japanese SDRT Corpus}
\label{sec:methods}
\subsection{Corpus Construction}
\label{Sec3.1}
To prepare training data for the Japanese SDRT parser, we translated MSDC and STAC into Japanese using GPT-5.1\footnote{
We used \texttt{gpt-5.1-2025-11-13}, \url{https://openai.com/index/gpt-5-system-card/}.
The prompt utilized for the translation is provided in Appendix~\ref{App:prompt_translation}.
}, a proprietary LLM accessed via the OpenAI API. 
We translated the corpora clause by clause in order to preserve the original order of the discourse units. 
Furthermore, in order to ensure context-aware and natural translations, we provided GPT-5.1 with the target clause and its corresponding dialogue session.

Because GPT-5.1 sometimes yielded inconsistent translations for EEUs in MSDC, we implemented a rule-based translation approach for all EEUs to ensure consistency. 
Specifically, we mapped predefined English keywords to their Japanese equivalents\footnote {
For more detailed discussion about this issue, please refer to Appendix~\ref{App: Translation inconsistencies for EEUs in MSDC}.
}.

In terms of STAC, we followed the methodology of \citet{llamipa} and applied several simplifications. 
Details are provided in Appendix~\ref{App:similification of the STAC}.

\subsection{Dataset Statistics}
The training set of MSDC contains 288 dialogues, 12,594 EDUs, and 3,439 EEUs, and the test set contains 97 dialogues, 3,886 EDUs, and 1,048 EEUs.
In MSDC, the relation \textit{Result} is the most frequent, followed by \textit{Acknowledgement}, \textit{Elaboration}, and \textit{Narration}.
The training set of STAC contains 1,017 dialogues, 11,380 EDUs, and 2,488 EEUs, and the test set contains 113 dialogues, 1,208 EDUs, and 257 EEUs\footnote{
For both corpora, the statistics are identical for the original English and translated Japanese versions.
}.
In STAC, \textit{Question-answer pair} is the most frequent relation, followed by \textit{Comment}, \textit{Acknowledgement}, and \textit{Result}.

\subsection{Quality Analysis}
We manually evaluated a sample of the Japanese MSDC corpus, our primary dataset, and automatically evaluated the entire MSDC corpus.

We performed two types of manual evaluation.
First, one of the authors, a native Japanese speaker, verified the naturalness of the translations and the consistency of the relation labels after translation.
Second, three of the authors, all native Japanese speakers, manually scored each speaker's turn on a scale from 0 to 100 based on translation quality. 
We randomly sampled and evaluated 20 dialogue sessions out of all 97 test sessions in the Japanese MSDC.
The mean score and standard error for each annotator are shown in Table~\ref{tab:trans_eval_manu}.
Inter-annotator agreement, measured by Krippendorff's $\alpha$ with the interval metric, was 0.132.
Although this value is low, limited agreement is a well-known challenge in human evaluation of machine translation~\citep{continuous_assessment}.
Examples of translation errors are presented in Appendix~\ref{App:error analysis of translation}.
\input{tables/trans_eval_manu}

For the automatic evaluation, we used XCOMET-XL~\citep{xcomet}.
XCOMET-XL can take the source and target texts and output a quality score ranging from 0.0 to 1.0, where higher scores indicate better translation quality.
We calculated scores for both the training and test sets of the Japanese MSDC.
The resulting XCOMET-XL scores were 0.84 for the training set, 0.85 for the test set. 

\input{tables/ja_llamipa_prompt}
\section{Japanese SDRT Parsers}
\subsection{Experimental Settings}
\paragraph{Llamipa}
We fine-tune two LLMs, \texttt{Llama-3-8B} and \texttt{Llama-3-Swallow-8B-v0.1} on our Japanese corpora, yielding two Japanese parsers.
Regarding the English parsers, we fine-tune the \texttt{Llama-3-8B} model on the original English corpus.
The Swallow model is based on Llama-3 and has been continually pre-trained on Japanese text\footnote{
Detailed settings are shown in Table~\ref{tab:model-details} in Appendix~\ref{App:hyperparameters}.}.
For the evaluation, we employ a predicted structure setting~(cf. Section \ref{SDRT parsing with LLMs}). %

\paragraph{Prompt-Based Parsers}
We compare Llamipa against several prompt-based parsers, evaluating each under zero-shot and few-shot (one-shot and four-shot) settings.
Our prompt design follows the proposal by \citet{fan-etal-2024-uncovering_promtmethod1}. 
In the original dataset, each dialogue corresponds to the entire process of building one structure in the \textit{Minecraft} game world. 
We count one such complete dialogue as a single ``shot'' and sample these exemplars from the training set.
We use GPT-5.5\footnote{
\texttt{gpt-5.5-2026-04-23}, \url{https://openai.com/index/gpt-5-5-system-card/}
} and \texttt{Llama-3.1-8B-Instruct} for English and Japanese experiments, along with \texttt{Llama-3.1-Swallow-8B-Instruct-v0.3} for the Japanese experiments.
Note that for prompt-based parsers, we adopt newer LLMs than those used for Llamipa because the context window of Llama-3 (8K tokens) is too short for these prompts.

\paragraph{Evaluation Metrics}
We calculate the F1-score for each discourse relation.
In order to evaluate the overall performance, we use the weighted average F1-score, where the F1-score for each discourse relation is weighted by its support (i.e., its frequency in the dataset).

\subsection{Results}
\input{tables/p_llamipa_STAC_MSDC}
We present the performance of Llamipa and prompt-based parsers on the Japanese MSDC in Table~\ref{ja_llamipa_llama_swallow_gpt}\footnote{
The performance for English is available in Appendix~\ref {App:detailed_results}.
}.
Overall, Llamipa achieves the numerically highest weighted average F1-score, which is around 0.79.
The weighted average F1-scores of prompt-based parsers do not exceed 0.72.

For individual relations, Llamipa outperforms prompt-based parsers for the majority of the relations.
However, prompt-based parsers outperform Llamipa for \textit{Conditional} and \textit{Explanation} relations.
These two relations have only a few instances in the training data.
This sparsity might have hindered the fine-tuning quality, making it difficult for Llamipa to learn these two relations.

Among the prompt-based parsers,
GPT-5.5 benefits from the increasing number of few-shot exemplars.
Its performance improves as the number of few-shot exemplars increases. 
In contrast, Llama-3.1 and Swallow-3.1 do not show the same trend.
For both models, the weighted average F1-score under the one-shot setting~(0.32 for Llama-3.1 and 0.30 for Swallow-3.1) is higher than that under zero-shot setting~(0.24 for Llama-3.1 and 0.22 for Swallow-3.1).
Nevertheless, for the four-shot setting, the weighted average F1-scores increase only minimally for Llama-3.1~(0.33), or even decrease substantially for Swallow-3.1~(0.09).
 
This is likely because adding more exemplars also increases the input length. 
Compared with GPT-5.5, which has a context window of roughly one million tokens, Llama-3.1 and Swallow-3.1 are both based on architectures that support context windows of up to 128K tokens. 
Specifically, the publicly released Swallow-3.1 checkpoint is configured with a maximum sequence length of 8,192 tokens, and the prompt length for the four-shot setting exceeds this, although the inputs were not truncated.
In contrast, GPT-5.5 might be better able to utilize information across long contexts and thus benefit more from additional exemplars.

We now turn to the best-performing parser, Llamipa, and compare its performance on the English and Japanese corpora.
Table~\ref{p_llamipa_STAC_MSDC} shows the SDRT parsing performance of Llamipa on the MSDC and STAC corpora.\footnote{We also calculate precision and recall for each relation. The results are presented in Appendix~\ref{App:detailed_results}.}
For both MSDC and STAC, performance on the original English corpus and the translated Japanese corpus is comparable.
Across all parsers, the weighted average F1-score is around 0.79 for MSDC and 0.60 for STAC.
Notably, the weighted average F1-score of the English Llama-3-based model on STAC is lower than the 0.78 reported by \citet{llamipa}.
This might be due to differences in the experimental setup, including 
computational environments and simplification of training data.

\section{Ablation Tests}
We have hypothesized that parsers might achieve different levels of performance.
However, as we have shown in the previous section, their performance is comparable.
To investigate why, we study which information the models rely on to identify discourse relations. 
We conduct three ablation analyses.
Our first study is about Japanese discourse markers.
Japanese discourse markers, such as sentence-final particles, are linguistic expressions that help speakers and listeners understand how utterances are related in discourse.
We evaluate the changes in performance before and after removing discourse markers from the input.

Our second and third studies are about two types of information in the input.
The input of Llamipa consists of three parts: the context clauses, the discourse relations in the context, and the target clause.
The context clauses are largely different in English and Japanese.
However, the discourse relations in the context are the same for both languages.
A plausible hypothesis for the comparable performance in English and Japanese is that parsers rely more on the discourse relations than on the context clauses.
We therefore ablate the discourse relations and the context clauses to verify this hypothesis.

\input{tables/statistics_ab}
\subsection{Settings}
\paragraph{Discourse Markers}
We select connectives that appear frequently~(at least 40 times) at the beginning of EDUs in the MSDC test data. 
This yields five connectives: \textit{sorekara} (それから; ``and then''), \textit{sositara} (そしたら; ``then''), \textit{jaa} (じゃあ; ``well/then''), \textit{dakara} (だから; ``so/therefore''), and \textit{demo} (でも; ``but'').

In the Japanese MSDC corpus, \textit{sorekara} frequently co-occurs with the discourse relation \textit{Narration}.
We hypothesize that the parser uses this expression as a strong cue when predicting \textit{Narration}.
Similarly, \textit{sositara} and \textit{jaa} often co-occur with \textit{Result} and \textit{Narration}, \textit{dakara} with \textit{Elaboration}, and \textit{demo} with \textit{Contrast}; we hypothesize that these expressions are crucial for predicting the discourse relations with which they most frequently co-occur.

Apart from the five connectives mentioned above, we also include the sentence-final particles \textit{yo}, \textit{ne}, and \textit{ka} according to the established theoretical discussion.
The particle \textit{yo} draws the listener's attention to information they are presumed not to know~\citep{BN04010577}.
The particle \textit{ne} confirms information that is assumed to be shared between the speaker and the listener~\citep{BN04010577}.
Based on these functions, we expect \textit{yo} to be crucial for \textit{Correction} relations and \textit{ne} to be crucial for \textit{Acknowledgement}.
The particle \textit{ka} expresses questions or doubts, so we expect it to be crucial for \textit{Clarification-Q}, \textit{Confirmation-Q}, and \textit{Question-answer pair}.
Table~\ref{stat_ab} presents the statistics of these expressions, relations, and their co-occurrences.

\paragraph{Contextual Discourse Relations}
For the discourse relations, we simply compare the parsers' performance with and without discourse relations given.
We conduct an ablation analysis by removing them all from the preceding context. 
Specifically, we provide the input context without any discourse relation information.

\paragraph{Context Clauses}
In order to investigate model performance when provided with only the discourse relations and the target clause, we mask the content of the context clauses.
We replace each context clause with the string \texttt{[MASKED]} while leaving other information unchanged.
An example of this process is shown in Figure~\ref{fig:mask_ex}.

For the aforementioned three studies, we use the Llama-3-based Llamipa parser for English and Japanese, and the Llama-3-Swallow-based Llamipa parser for Japanese only.
These studies are conducted on the MSDC test set and under the gold structure setting~(cf. Section~\ref{SDRT parsing with LLMs}) in order to keep the contextual discourse relations of the input fixed.
Under the predicted structure setting, the input itself could differ before and after each ablation, making it difficult to isolate the effect of the ablated component.
\input{figures/mask_ex}
\input{tables/ablation}

\input{tables/ab_context}
\input{tables/ab_mask}
\subsection{Results}
Table \ref{ablation} presents the results of the ablation tests for the discourse markers~(connectives and sentence-final particles). 
We compare the F1-scores before and after ablation and measure the number of initial predictions that remained unchanged after the target expressions were removed.

When \textit{demo} is removed, the F1-score for predicting \textit{Contrast} decreases from 0.73 to 0.11 for the Llama-3-based model and from 0.77 to 0.14 for the Llama-3-Swallow-based model.
Furthermore, of the 46 \textit{Contrast} predictions in which \textit{demo} appears in the second argument, only 2 remained unchanged after ablation for the Llama-3-based model (2 of 48 for the Llama-3-Swallow-based model).
These results suggest that the constructed Japanese SDRT parsers rely heavily on the presence of \textit{demo} to identify \textit{Contrast}.

In contrast, removing \textit{sorekara} changed a considerable proportion of the \textit{Narration} predictions: 15 of the 46 predictions changed after ablation for the Llama-3-based model (10 of 43 for the Llama-3-Swallow-based model). 
Nevertheless, the F1-score for \textit{Narration} remained almost unchanged (0.89 $\rightarrow$ 0.90 for both models), possibly because these instances account for only a small fraction of the \textit{Narration} relations in the test set. 
This suggests that the models use \textit{sorekara} as a cue for predicting \textit{Narration}, although its impact on overall performance is limited.

However, for the other connectives (\textit{sositara}, \textit{jaa}, and \textit{dakara}), and the sentence-final particles (\textit{yo}, \textit{ne},
and \textit{ka}), no substantial changes in performance are observed when predicting their respective discourse relations. 
This suggests that our hypothesis that these expressions contribute to predicting specific discourse relations was not empirically supported.

Table~\ref{ab_context} presents the results of the ablation study on the gold discourse relations among the context clauses.
The performance substantially decreases; the weighted average F1-scores drop by at least 0.29 for all parsers.
The largest decrease is observed for \textit{Narration}, which drops by at least 0.85.
Notably, the decrease for \textit{Contrast} in the Japanese models is relatively small.
This is consistent with the result of the first ablation, which suggests that the models rely heavily on \textit{demo}.
Since \textit{demo} appears in the target clause, it remains available to the models even after the contextual discourse relations are removed.

Table~\ref{ab_mask_context} presents the performance of the models before and after masking the content of the context clauses.
Across all models, we only observe a slight decline in the weighted average.
Combined with the results for the ablation on gold discourse relations in input, these results suggest that the parsers primarily rely on the discourse relations in the input rather than on the context clauses, supporting our hypothesis.
Thus, the reason why Japanese and English SDRT parsers show comparable performance might be that parsers rely on the discourse structure, which is shared across Japanese and English corpora, rather than on the content of the context clauses.

\section{Conclusion}
In this paper, we used Japanese as a test case to investigate the cross-linguistic generalizability of LLM-based Segmented Discourse Representation Theory (SDRT) parsing.
We first constructed Japanese corpora by automatically translating the English MSDC and STAC corpora.
We then compared the performance of fine-tuning-based and prompt-based parsing approaches on Japanese and English.
Our results show that fine-tuned parsers achieve higher SDRT parsing performance than prompt-based parsers and perform comparably across English and Japanese.

In order to explain why English and Japanese models perform comparably, we conducted three ablation studies to analyze the types of information on which the models rely when identifying discourse relations.
First, we studied the influence of Japanese connectives and sentence-final particles that frequently co-occur with particular discourse relations in our translated Japanese MSDC corpus.
The results show that \textit{demo} (でも; ``but'') particularly influences the prediction of the \textit{Contrast} relation.
However, the impact of the other expressions was not as large as that of \textit{demo}.
Our second and third ablation studies examined the effects of removing the gold discourse relations among the context clauses and of masking the context clauses in the input, respectively.
The results suggest that the models rely heavily on the discourse relations provided in the input.
This reliance may explain the small performance difference between our English and Japanese SDRT parsers.

Overall, our findings suggest the cross-linguistic generalizability of LLM-based SDRT parsing to Japanese and show that language-specific cues, such as \textit{demo}, can facilitate the prediction of particular discourse relations. 
We achieved strong Japanese discourse parsing performance using a simple approach based on clause-by-clause translation and model training, suggesting that generative LLMs offer considerable potential for cross-lingual transfer in discourse parsing.

Building on these results, one promising direction for future work is to investigate the generalizability of LLM-based discourse parsing to languages beyond English and Japanese. Such an investigation is crucial for extending discourse parsing to a wider range of languages.
Another worthwhile direction is to conduct more fine-grained analyses that distinguish linguistic expressions providing critical cues for discourse-relation prediction from those that provide little or no useful information. Such analyses may clarify the discourse functions of specific lexical items in Japanese and ultimately improve Japanese discourse parsing performance.
Finally, a promising direction for future work is to analyze how LLMs process discourse relations using interpretability methods such as that of \citet{miao-kan-2025-discursive}, which would shed light on how LLMs represent and interpret discourse structure.

\section*{Limitations}
In this study, we used only MSDC and STAC, which are corpora of chat logs collected during gaming sessions.
Moreover, due to time and staffing constraints, we did not manually evaluate the entire Japanese translation corpus.
One direction for future research is to expand such resources to include other domains and genres.
Another issue regarding corpora is that the Japanese corpora are constructed based on machine translation.
Translated Japanese might be unnatural and different from what native Japanese speakers produce.
This unnaturalness and difference might introduce confounding factors that affect parser performance.
Therefore, constructing a native Japanese SDRT corpus is also necessary as future work. 
\section*{Ethical Statement}
In this study, we construct and release a Japanese Segmented Discourse Representation Theory (SDRT) corpus by translating existing English corpora (the MSDC and STAC datasets) using a Large Language Model (LLM). This dataset is released strictly for academic and research purposes. In doing so, we carefully consider all licensing and copyright implications. Because our Japanese corpus is derived from the original English datasets, we strictly adhere to their terms of use and licenses. The translated dataset will be distributed under an appropriate license that aligns with the restrictions of the source materials, ensuring that the original authors' intellectual property is fully respected.

\section*{Acknowledgements}
We thank the lab members for their insightful advice and the anonymous reviewers for their helpful comments and feedback.
This work was supported by JSPS KAKENHI Grant Number JP24H00809, Japan.

\bibliography{custom}

\newpage
\appendix

\section{Semantics of Discourse Relations used in MSDC and STAC}
\label{App:semantics}
Below are the relations used in MSDC and STAC, along with illustrative examples of each.
In the dataset, relation instances are represented as pairs of indices associated with a specific discourse relation label.
The first and second arguments $(x,y)$ represent the source and target of the directed edge in the discourse graph, respectively.
In the following examples, the upper clause corresponds to argument $x$ and the lower clause to argument $y$.
\begin{itemize}
    \item \textit{Acknowledgement}$(x,y)$...$y$ entails that the speaker of $y$ has accepted or achieved the intention of the speaker of $x$
    \begin{exe}
    \ex
    \begin{xlist}
    \ex Builder: ``place orange 2 1 2, place orange 0 1 3''
    \ex Architect: ``Great!''
    \end{xlist}
    \end{exe}
    \item \textit{Alternation}$(x,y)$...Either $x$ or $y$ is true.
    \begin{exe}
        \ex
        \begin{xlist}
            \ex Builder: ``so parallel?''
            \ex Builder: ``or perpendicular?''
        \end{xlist}
    \end{exe}
    \item \textit{Background}$(x,y)$...$x$ provides some stage setting for the event that happens in $y$, or vice versa.
    \begin{exe}
        \ex
        \begin{xlist}
            \ex Dave: While we're talking about trading,
            \ex Dave: does anyone have sheep?
        \end{xlist}
    \end{exe}
    \item \textit{Clarification-Q}$(x,y)$...$y$ provides a question about $x$. It occurs when one does not follow what happened or what is said.
    \begin{exe}
        \ex
        \begin{xlist}
            \ex Architect: ``anyway please place the first purple block on the rightmost red block''
            \ex Builder: ``is it connected?''
        \end{xlist}
    \end{exe}
    \item \textit{Comment}$(x,y)$...$y$ provides an opinion or evaluation of the content associated with $x$.
    \begin{exe}
        \ex
        \begin{xlist}
            \ex Architect: ``I am sorry for the incorrect instructions''
            \ex Builder: ``you did a good job!''
        \end{xlist}
    \end{exe}
    \item \textit{Conditional}$(x,y)$...$x$ is a hypothesis and $y$ is a consequence of the hypothesis.
    \begin{exe}
        \ex
        \begin{xlist}
            \ex Architect: ``so if you take a directly top-down view,''
            \ex Architect: ``it will look like a staircase going up and right''
        \end{xlist}
    \end{exe}
    \item \textit{Confirmation-Q}$(x,y)$...$y$ is a question about whether $x$ is correct.
    \begin{exe}
        \ex
        \begin{xlist}
            \ex Builder: ``place orange -1 4 1, place orange -1 4 0, place orange -1 4 -1''
            \ex Builder: ``like that?''
        \end{xlist}
    \end{exe}
    \item \textit{Continuation}$(x,y)$...Both $x$ and $y$ are subject to the same topic.
    \begin{exe}
        \ex
        \begin{xlist}
            \ex Architect: ``you will be removing the middle block of the last row you added.''
            \ex Architect: ``and the middle three blocks of the row youre about to add.''
        \end{xlist}
    \end{exe}
    \item \textit{Contrast}$(x,y)$...$x$ and $y$ have similar semantic structures, but contrasting themes.
    \begin{exe}
        \ex
        \begin{xlist}
            \ex Architect: ``I have bad news''
            \ex Architect: ``but it's an easy fix''
        \end{xlist}
    \end{exe}
    \item \textit{Correction}$(x,y)$...$y$ corrects a part of $x$.
    \begin{exe}
        \ex
        \begin{xlist}
            \ex Architect: ``and 1 space to the right''
            \ex Architect: ``I meant to say 1 space to the left''
        \end{xlist}
    \end{exe}
    \item \textit{Elaboration}$(x,y)$...$y$ provides more information about the eventuality introduced in $x$.
    \begin{exe}
        \ex
        \begin{xlist}
            \ex Architect: ``we'll now do orange''
            \ex Architect: ``place on orange to the right of the red closest to you''
        \end{xlist}
    \end{exe}
    \item \textit{Explanation}$(x,y)$...$y$ explains why, or gives the cause of, what happened in $x$.
    \begin{exe}
        \ex
        \begin{xlist}
            \ex Builder: ``I'll stay in a fixed position''
            \ex Builder: ``so it's easier to give me directions with respect to what I'm looking at''
        \end{xlist}
    \end{exe}
    \item \textit{Narration}$(x,y)$...The main eventualities of $x$ and $y$ occur in the sequence in which they are described.
    \begin{exe}
        \ex
        \begin{xlist}
            \ex Architect: ``then 1 red on top of the right red''
            \ex Architect: ``then another red to the bottom right of the right ground red block.''
        \end{xlist}
    \end{exe}
    \item \textit{Parallel}$(x,y)$...$x$ and $y$ share a common topic and have similar semantic structures.
    \begin{exe}
        \ex
        \begin{xlist}
            \ex You can trade for more sheep with Rennoc,
            \ex You can also trade with me
        \end{xlist}
    \end{exe}
    \item \textit{Question-answer pair}$(x,y)$...$x$ is a question, and $y$ is an answer to it.
    \begin{exe}
        \ex
        \begin{xlist}
            \ex Builder: ``Whats our mission this time?''
            \ex Architect: ``we have a difficult one ahead of us this time build\-er.''
        \end{xlist}
    \end{exe}
    \item \textit{Q-Elaboration}$(x,y)$...$y$ is a question intended to get more information ab\-out $x$.
    \begin{exe}
        \ex
        \begin{xlist}
            \ex Architect: ``we have a difficult one ahead of us this time builder.''
            \ex  Builder: ``describe the structure?''
        \end{xlist}
    \end{exe}
    \item \textit{Result}$(x,y)$...The main eventuality of $x$ is understood to cause the eventuality given by $y$.
    \begin{exe}
        \ex
        \begin{xlist}
            \ex Architect: ``now connect these red blocks with three more red blocks making a 3x3''
            \ex Builder: ``place red 0 7 0, place red 0 6 0, place red 0 5 0''
        \end{xlist}
    \end{exe}
    \item \textit{Sequence}$(x,y)$...$x$ and $y$ happen consecutively.    
    \begin{exe}
        \ex
        \begin{xlist}
            \ex Builder: ``Mission has started''
            \ex Builder: ``place purple -1 1 3, pick purple -1 1 3''
        \end{xlist}
    \end{exe}
\end{itemize}

\section{Translation Examples and Error Analysis for MSDC}
\label{App:error analysis of translation}
In this section, we provide examples of translations and several error analyses of MSDC.
Table~\ref{translations} shows examples of the translations. The sentence above is the original English, while the sentence below is the Japanese translation.

Examples of translation errors are presented in Table~\ref{error}.
In each example, the upper segment provides the original English excerpt, while the lower segment presents its corresponding Japanese translation.

The first example is shown in the upper section of the table.
In this excerpt, the Builder intends to ask the Architect if any color of block can be placed, saying ``any color.''
However, the resulting translation fails to reflect the interrogative intent of the original utterance.
Therefore, a more appropriate translation would be ``どんな色でもいい？'' (``Is any color fine?'') rather than ``どんな色でもいいよ'' (``Any color is fine.'').

The second excerpt is located in the middle section of the table.
In this instance, the translation of the preposition ``on'' is incorrect.
The speaker intends for the Builder to place a block to the left of the topmost block.
Consequently, the translation ``一番上のオレンジのブロックの上に'' (``on top of the top orange block'') is inappropriate.
The correct translation should be ``一番上のオレンジのブロックに''(``on the top orange block'').

The final example is located in the bottom section of the table.
In this case, the phrase ``上と下に'' (``on top and bottom'') is redundantly repeated.
Therefore, the phrase ``上と下に'' in the second clause should be removed to ensure a more natural translation.

\input{tables/translation_ex}
\input{tables/error_ex}

\section{Detailed Experimental Results}
\label{App:detailed_results}
Tables~\ref{result_all_msdc} and \ref{result_all_STAC} present the detailed performance metrics (precision, recall, and F1-score) of the constructed SDRT parsers on the MSDC and STAC corpora, respectively. 
Table~\ref{eval_sum_subset} details the performance of the prompt-based methods on the MSDC corpus in both English and Japanese.

\section{Simplification of STAC}
\label{App:similification of the STAC}
We applied several simplifications to the dataset following the methodology of \citet{llamipa} because Llamipa cannot deal with CDUs directly.
First, we flattened CDUs into their constituent EDUs or EEUs. 
Second, we merged consecutive EEUs into single units. 
Finally, we removed any EEUs that lacked relations to EDUs.

\section{Evaluation on All Relations of MSDC}
\label{App: Evaluation on All Relation of MSDC}
In Tables~\ref{ja_llamipa_llama_swallow_gpt} and~\ref{p_llamipa_STAC_MSDC}, some gold relations in the original MSDC dataset are excluded from the evaluation following \citet{llamipa}. 
This is because Llamipa does not predict relations whose distance exceeds its context window size. 
Therefore, such relations are excluded from the training and test data. 
The evaluation for all relations is presented in Table~\ref{eval_all_rel}.

\section{LLM-translation inconsistencies for EEUs in MSDC}
\label{App: Translation inconsistencies for EEUs in MSDC}
Directly translating EEUs using GPT-5.1 frequently introduces inconsistencies.
The following examples illustrate this issue:

\begin{exe} 
\ex\label{notranslate} 
\begin{xlist} 
\ex place green 0 3 -1, place green 0 4 -1 
\ex place green 0 3 -1, place green 0 4 -1
\end{xlist} 
\ex\label{translated} 
\begin{xlist} 
\ex pick red 2 7 5
\ex 赤を 2 7 5 から取る 
\end{xlist}
\ex\label{ill_translation} 
\begin{xlist}
\ex pick purple -2 4 1, pick purple 1 4 1 
\ex 紫を -2 4 1 を回収、紫を 1 4 1 を回収 
\end{xlist} 
\end{exe}

In each example, the first line (a) represents the original English EEU, and the second line (b) shows the LLM-generated translation.
In Example (\ref{notranslate}), the EEU remains completely untranslated, whereas in Example (\ref{translated}), it is translated into Japanese. 
Furthermore, the translation in Example (\ref{ill_translation}) is ungrammatical. 

Because such inconsistencies could negatively impact the performance of SDRT parsers, we implemented a rule-based translation approach for all EEUs to ensure structural and linguistic consistency.
Specifically, ``place'' was mapped to ``設置'' and ``pick'' to ``採取''.
The color words ``orange,'' ``blue,'' ``green,'' ``yellow,'' ``red,'' and ``purple'' were mapped to ``オレンジ,'' ``青,'' ``緑,'' ``黄色,'' ``赤,'' and ``紫,'' respectively.

\section{The prompt used for the translation}
\label{App:prompt_translation}
This appendix provides the prompts utilized to translate the original MSDC and STAC into Japanese.
The prompts below are reproduced as used in our experiments, including typographical errors in the original prompts (e.g., "Ouptput" for "Output" and "as possible as naturally" for "as naturally as
possible").
We retain them so that our experiments can be replicated exactly.
The following prompt is used for MSDC.
Note that although the MSDC prompt mentions translations of the two previous clauses, these were not included in the actual input; the model received only the dialogue session and the target clause, as described in Section~\ref{Sec3.1}.
\newpage
\begin{tcolorbox}[breakable]
\small
\begin{Verbatim}[
  breaklines=true,              %
  breakanywhere=true,           %
  breaksymbolleft={},
  breakindent=2em,
  xleftmargin=1em               %
]
{"role": "system",
"content":
"""
## Task
You are given a multi-turn Minecraft chat log in JSON format,a target clause,
and your translation of the two previous clauses of the target clause.
The target clause is one of the clauses in the chat log.
Translate the clause into Japanese as possible as naturally. 

## INPUT FORMAT
You are given the following JSON structure as input: 

```
# Context
[{"text": "<clause1>","speaker": "<speaker1>"},{"text": "<clause2>","speaker": "<speaker2>"},...]
# Target clause
{"text": "<target_clause>","speaker": "<target_speaker>"}
```

## Rules
- No comments
- Translate only the target clause
- Ouptput only the translated text
- The transcript includes descriptions of in-game actions. Keep them as they are.
    - For example, if the text is "pick yellow 0 0 1, place red 1 0 0", 
    Do not translate the action commands. Keep them in English
- Translate as naturally as possible while considering the context and previous translated clauses.
"""
}
{"role": "user", "content": "#Context\n" + <dialogue session>}
{"role": "user", "content": "# Target clause\n" + <target clause>}
\end{Verbatim}
\end{tcolorbox}

The following is used for STAC.
\begin{tcolorbox}[breakable]
\small
\begin{Verbatim}[
  breaklines=true,              %
  breakanywhere=true,           %
  breaksymbolleft={},
  breakindent=2em,
  xleftmargin=1em               %
]
{"role": "system",
"content":
"""
## Task
You are an expert translator specializing in board games.
Your task is to translate logs from the online version of the game The Settlers of Catan (STAC corpus) into natural Japanese.
You will be provided with a target clause and the surrounding dialogue context.

## INPUT FORMAT
```
# Context
[{"text": "<clause1>","speaker": "<speaker1>"},{"text": "<clause2>","speaker": "<speaker2>"},...]

# Target clause
{"text": "<target_clause>","speaker": "<target_speaker>"}
```

## Rules
- Translate only the target clause.
- Output only the translated text. Do not include any explanations or comments.
- Tone & Style:
    - If the speaker is a player: Use a natural, conversational chat style.
    - If the speaker is "Server" or "UI": These represent system actions or game events. Translate them as clear, concise system messages or objective descriptions.
- Ensure the translation reflects the strategic intent of the negotiation if the context involves trading.
- Do not translate player names. Keep them exactly as they appear in the original text.
"""
}
{"role": "user", "content": "#Context\n" + <dialogue session>}
{"role": "user", "content": "# Target clause\n" + <target clause>}   
\end{Verbatim}
\end{tcolorbox}

\section{Prompts for Prompt-Based Discourse Parsing}
\label{App:prompts}
This appendix presents the full text of the prompts utilized in our prompt-based experiments. 
The system prompt consists of general instructions and examples for in-context learning (ICL), while the user prompt contains the target dialogue for discourse parsing.
The prompt is reproduced as used in our experiments. 
Note that it contains errors, including "Segmented Discourse Rhetorical Theory" (for "Segmented Discourse Representation Theory") and "Dialougue" (for "Dialogue"). 
We retain them because all reported results were obtained with this prompt.
\newpage
\begin{tcolorbox}[breakable]
\small
\begin{Verbatim}[
  breaklines=true,              %
  breakanywhere=true,           %
  breaksymbolleft={},
  breakindent=2em,
  xleftmargin=1em               %
]
{"role": "system",
"content": 
According to the Segmented Discourse Rhetorical Theory, the rhetorical structure of a dialogue can be represented by a directed acyclic graph, where nodes are utterances or actions and edges are the following 16 relations:
["Elaboration", "Acknowledgement", "Continuation", "Contrast", "Correction", "Result", "Comment", "Question-answer_pair", "Narration", "Clarification_question", "Confirmation_question", "Q-Elab", "Explanation", "Alternation", "Conditional", "Sequence"]
please annotate the rhetorical structure of the following dialogue and represent it in the form of [index1, index2, 'relation'], where index1 and index2 are the index of two utterances, and the 'relation' is one of the above relations to connect the two utterances.

--------------------
Dialougue:
0: Builder said, Mission has started .
1: Builder said, hi
...
Annotation:
[
[0, 1, "Acknowledgement"],
...
]
--------------------
}
{"role": "user", 
"content":
"0: Builder said, Mission has started.
1: Architect said, We'll be making a pair of horseshoes
2: Architect said, First, start by making a 3x3 square of red bricks on the ground
...
"}
\end{Verbatim}
\end{tcolorbox}

\newpage
\section{Hyperparameters}
Table~\ref{tab:model-details} details the hyperparameters used for training the SDRT parsers.
\label{App:hyperparameters}
\input{tables/hyperparameters}

\input{tables/results_all_MSDC}
\input{tables/results_all_STAC}
\input{tables/eval_summary_subset_llamipa}
\input{tables/eval_all_rel}
\end{document}

%% file: figures/MSDC_figure.tex
\begin{figure}[t]
    \centering
    \includegraphics[width=\linewidth]{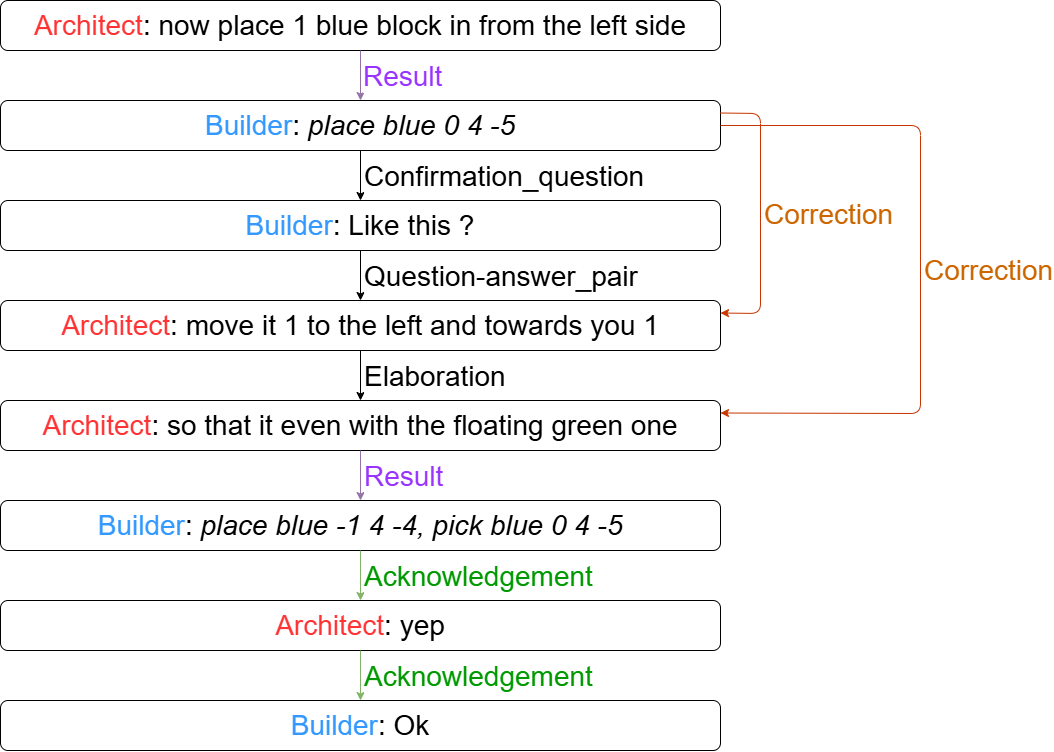}
    \caption{An extract from MSDC, represented as a discourse graph. EDUs and EEUs are shown in roman and italic fonts, respectively.}
    \label{fig:MSDC}
\end{figure}

%% file: tables/trans_eval_manu.tex
\begin{table}[h]
    \centering
    \small
    \begin{tabular}{@{}llll@{}}
    \toprule
                    & Annotator1 & Annotator2 & Annotator3 \\ \midrule
    Mean            & 97.2      & 88.3      & 88.7\\
    Standard error  & 0.3 & 0.2 & 0.3\\
    \bottomrule
    \end{tabular}
    \caption{Mean translation evaluation scores and standard errors for each annotator. }
    \label{tab:trans_eval_manu}
\end{table}

%% file: tables/ja_llamipa_prompt.tex
\begin{table*}[]
\centering
\resizebox{\textwidth}{!}{
\begin{tabular}{@{}lcccccccccccc@{}}
\toprule
Model            & Llama-3        & Swallow-3      & \multicolumn{3}{c}{Llama-3.1} & \multicolumn{3}{c}{Swallow-3.1} & \multicolumn{3}{c}{GPT-5.5} &          \\ \cmidrule(lr){2-2} \cmidrule(lr){3-3} \cmidrule(lr){4-6} \cmidrule(lr){7-9}\cmidrule(lr){10-12}
Method           & \multicolumn{2}{c}{Llamipa}   & 0shot    & 1shot   & 4shot   & 0shot    & 1shot    & 4shot    & 0shot   & 1shot   & 4shot  & Supports \\ \midrule
Ackn.            & \textbf{0.86} & 0.85          & 0.56     & 0.43    & 0.43    & 0.51     & 0.36     & 0.15    & 0.56    & 0.76    & 0.78   & 792      \\
Alte.            & 0.81          & \textbf{0.90} & 0.00     & 0.00    & 0.42    & 0.00     & 0.00     & 0.13     & 0.76    & 0.65    & 0.88   & 24       \\
Clari-Q.         & 0.67          & \textbf{0.68} & 0.08     & 0.24    & 0.27    & 0.01     & 0.19     & 0.08     & 0.40    & 0.58    & 0.60   & 134      \\
Comm.            & \textbf{0.58} & \textbf{0.58} & 0.16     & 0.23    & 0.12    & 0.19     & 0.14     & 0.03     & 0.34    & 0.47    & 0.49   & 382      \\
Cond.            & 0.00          & 0.00          & 0.00     & 0.00    & 0.00    & 0.00     & 0.00     & 0.00     & \textbf{0.22}    & \textbf{0.22}    & 0.00   & 13       \\
Conf-Q.           & \textbf{0.91}          & 0.90          & 0.07     & 0.56    & 0.56    & 0.01     & 0.55     & 0.14     & 0.78    & 0.89    & 0.90   & 159      \\
Contin.          & \textbf{0.45} & 0.42          & 0.04     & 0.06    & 0.07    & 0.06     & 0.04     & 0.03    & 0.11    & 0.26    & 0.40   & 321      \\
Contr.           & 0.76          & \textbf{0.81} & 0.03     & 0.10    & 0.07    & 0.04     & 0.10     & 0.00     & 0.55    & 0.77    & 0.73   & 68       \\
Corr.            & \textbf{0.75} & 0.74          & 0.03     & 0.08    & 0.07    & 0.03     & 0.08     & 0.01     & 0.22    & 0.49    & 0.60   & 403      \\
Elab.            & \textbf{0.76}          & 0.75          & 0.32     & 0.37    & 0.41    & 0.30     & 0.38     & 0.12    & 0.55    & 0.72    & 0.73   & 720      \\
Expl.            & 0.00          & 0.00          & 0.02     & 0.04    & 0.05    & 0.01     & 0.04     & 0.00     & 0.10    & 0.15    & \textbf{0.16}   & 16       \\
Narr.            & 0.80          & \textbf{0.81}          & 0.00     & 0.00    & 0.01    & 0.00     & 0.00     & 0.00     & 0.00    & 0.17    & 0.68   & 693      \\
QAP.             & 0.81          & \textbf{0.82} & 0.03     & 0.41    & 0.46    & 0.03     & 0.37     & 0.11     & 0.63    & 0.78    & 0.79   & 290      \\
QElab.           & 0.39          & \textbf{0.42} & 0.00     & 0.00    & 0.00    & 0.00     & 0.00     & 0.00     & 0.10    & 0.12    & 0.17   & 38       \\
Resu.            & \textbf{0.90} & \textbf{0.90} & 0.35     & 0.50    & 0.52    & 0.30     & 0.51     & 0.12    & 0.51    & 0.73    & 0.85   & 1765     \\
Sequ.            & \textbf{0.29} & 0.00          & 0.00     & 0.00    & 0.00    & 0.00     & 0.00     & 0.00     & 0.00    & 0.00    & 0.00   & 6        \\ \midrule
Weighted Average & \textbf{0.79} & \textbf{0.79} & 0.24     & 0.32    & 0.33    & 0.22     & 0.30     & 0.09     & 0.41    & 0.60    & 0.72   &          \\ \bottomrule
\end{tabular}
}
\caption{\label{ja_llamipa_llama_swallow_gpt}F1-scores for Llamipa and prompt-based parsers on the Japanese MSDC. "Supports" represents the number of instances per relation in the test data, and discourse relations are abbreviated. Bold values indicate the highest F1-scores. The models were evaluated on a subset of gold relations from the MSDC corpus (cf. Appendix~\ref{App: Evaluation on All Relation of MSDC}).}
\end{table*}

%% file: tables/p_llamipa_STAC_MSDC.tex
\begin{table*}[h]
\centering
\small
\begin{tabular}{@{}lllllllll@{}}
\toprule
Dataset    & \multicolumn{4}{l}{MSDC}                                                            & \multicolumn{4}{l}{STAC}                                                            \\ \cmidrule(lr){2-5} \cmidrule{6-9} 
Language   & EN                     & JA                     & JA                     & Supports & EN                     & JA                     & JA                     & Supports \\
Base model & Llama-3                & Llama-3                & Swallow-3               &          & Llama-3                & Llama-3                & Swallow-3                &          \\ \midrule
Ackn.      & \textbf{0.86} & \textbf{0.86} & 0.85                   & 792      & 0.61                   & 0.62                   & \textbf{0.65} & 173      \\
Alte.      & \textbf{0.90} & 0.81                   & \textbf{0.90} & 24       & \textbf{0.67} & 0.59                   & 0.62                   & 10       \\
Back.       & n/a                    & n/a                    & n/a                    & n/a      & 0.00                   & \textbf{0.20}                   & 0.00                   & 9        \\
Clari-Q.   & \textbf{0.69} & 0.67                   & 0.68                   & 134      & \textbf{0.50} & 0.46                   & \textbf{0.50} & 48       \\
Comm.      & 0.58                   & 0.58                   & 0.58                   & 382      & 0.60                   & 0.60                   & 0.60                   & 222      \\
Cond.      & 0.00                   & 0.00                   & 0.00                   & 13       & 0.53                   & \textbf{0.78} & 0.63                   & 12       \\
Conf-Q.    & \textbf{0.93} & 0.91                   & 0.90                   & 159      & n/a                    & n/a                    & n/a                    & n/a      \\
Contin.    & \textbf{0.45} & \textbf{0.45} & 0.42                   & 321      & 0.29                   & 0.25                   & \textbf{0.36} & 69       \\
Contr.     & 0.76                   & 0.76                   & \textbf{0.81} & 68       & \textbf{0.50} & 0.43                   & 0.49                   & 60       \\
Corr.      & 0.72                   & \textbf{0.75} & 0.74                   & 403      & \textbf{0.38} & 0.24                   & 0.31                   & 23       \\
Elab.      & \textbf{0.77} & 0.76                   & 0.75                   & 720      & \textbf{0.51} & 0.46                   & 0.44                   & 100      \\
Expl.      & 0.00                   & 0.00                   & 0.00                   & 16       & \textbf{0.39} & 0.37                   & 0.38                   & 55       \\
Narr.      & \textbf{0.82} & 0.80                   & 0.81                   & 693      & 0.00                   & 0.00                   & 0.00                   & 8        \\
Para.       & n/a                    & n/a                    & n/a                    & n/a      & \textbf{0.34} & 0.29                   & \textbf{0.34} & 23       \\
QAP.       & 0.81                   & 0.81                   & \textbf{0.82} & 290      & \textbf{0.81} & \textbf{0.81} & 0.78                   & 273      \\
QElab.     & \textbf{0.43}                   & 0.39 & 0.42                   & 38       & 0.49                   & \textbf{0.53} & 0.50                   & 84       \\
Resu.      & 0.90 & 0.90                  & 0.90                 & 1765      & \textbf{0.65} & 0.61                   & 0.61                   & 145      \\
Sequ.      & 0.25 & \textbf{0.29} & 0.00                   & 6      & \textbf{0.78} & \textbf{0.78} & 0.75                   & 124      \\ \midrule
Weighted Average   & 0.79                   & 0.79                   & 0.79                   &          & \textbf{0.60}                   & 0.59                   & 0.59                   &          \\ \bottomrule
\end{tabular}
\caption{\label{p_llamipa_STAC_MSDC}F1-scores of Llamipa on the MSDC and STAC datasets. "Supports" indicates the number of instances per relation in the test data. Discourse relations are abbreviated. "n/a" indicates that the relation is not included in the respective dataset. Bold values denote the best F1-scores. }
\end{table*}

%% file: tables/statistics_ab.tex
\begin{table}[h]
\centering
\small
\begin{tabular}{@{}lllc@{}}
\toprule
Expression      & Relation             & Occurrence & Position \\ \midrule
それから (\textit{sorekara})                   & Narr.            & 36/54                   & $y$                              \\ \hline
\multirow{2}{*}{そしたら  (\textit{sositara})} & Narr.            & 36/40                   & $y$                              \\
                           & Resu.               & 35/40                   & $y$                              \\ \midrule
\multirow{2}{*}{じゃあ (\textit{jaa})}       & Narr.            & 284/379                 & $y$                             \\
                           & Resu.            & 316/379                 & $y$                              \\ \midrule
だから (\textit{dakara})                     & Elab.          & 41/47                   & $y$                              \\ \midrule
でも (\textit{demo})                    & Contr.             & 46/54                   & $y$                              \\ \midrule
よ (\textit{yo})                      & Corr.           & 22/222                  & $y$                              \\ \midrule
ね (\textit{ne})                         & Ackn.      & 145/369                 & $y$                              \\ \midrule
\multirow{3}{*}{か (\textit{ka})}        & Clari-Q.      & 24/60                   & $y$                              \\
                           & Conf-Q.      & 8/60                    & $y$                              \\
                           & QAP. & 37/60                   & $x$  \\ \bottomrule  
\end{tabular}
\caption{\label{stat_ab} The statistics of the ablation study on target expressions. The ``Occurrence'' column is presented in the format $(a / b)$, where $a$ is the number of times the linguistic expression appears at the position specified in the ``Position'' column, and $b$ is the total number of clauses in which the expression occurs. In the ``Position'' column, $x$ and $y$ denote the first and second arguments of the discourse relation, respectively.}
\end{table}

%% file: figures/mask_ex.tex
\begin{figure}[h]
    \centering
    \includegraphics[width=\linewidth]{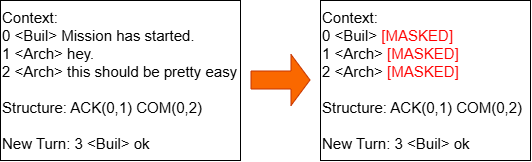}
    \caption{Depiction of the context clause masking.}
    \label{fig:mask_ex}
\end{figure}

%% file: tables/ablation.tex
\begin{table}[t]
\centering
\resizebox{\columnwidth}{!}{
\begin{tabular}{@{}cccccc@{}}
\toprule
Model                         &        & \multicolumn{2}{c}{Llama-3-base}   &\multicolumn{2}{c}{Swallow-base} \\ \cmidrule(l){3-4} \cmidrule(l){5-6}
Expression                   & Rel.   & F1 & Pred. & F1 & Pred. \\  \midrule
\makecell{それから \\ (\textit{sorekara})}                  & Narr.      & 0.89/0.90  & 46/31       &  0.89/0.90 & 43/33  \\ \midrule
\multirow{2}{*}{\makecell{そしたら \\ (\textit{sositara})}} & Narr.      & 0.89/0.89  & 35/33       &  0.89/0.89 & 36/34  \\
                              & Resu.      & 0.91/0.91  & 35/34  & 0.91/0.91    & 36/35    \\ \midrule
\multirow{2}{*}{\makecell{じゃあ \\ (\textit{jaa})}}      & Narr.      & 0.89/0.89 & 282/272     &  0.89/0.88  & 290/274    \\
                              & Resu.      & 0.91/0.91 & 308/300    & 0.91/0.91  & 313/298    \\ \midrule
\makecell{だから \\ (\textit{dakara})}                    & Elab.      & 0.77/0.77 & 42/41       &  0.76/0.76  & 39/39    \\ \midrule
でも (\textit{demo})                      & Contr.     & 0.73/0.11 & 46/2        &  0.77/0.14  & 48/2 \\ \midrule
よ (\textit{yo})                        & Corr.      & 0.79/0.79  & 30/28       &  0.81/0.81 & 22/20 \\ \midrule
ね (\textit{ne})                        & Ackn.      & 0.86/0.85 & 145/142     &  0.86/0.86  & 140/135\\ \midrule
\multirow{3}{*}{か (\textit{ka})}       & Clari-Q.    & 0.68/0.69  &23/23        &  0.68/0.67   & 26/24   \\
                              & Conf-Q.    & 0.92/0.92  & 5/5        & 0.91/0.91 & 4/4  \\
                              & QAP.       & 0.83/0.83  & 35/30      & 0.85/0.85 & 34/33  \\ \bottomrule
\end{tabular}
}
\caption{\label{ablation}F1-scores and the number of predictions (Pred.) before and after ablation. In the ``F1'' column, the notation $b/a$ indicates the scores before ($b$) and after ($a$) ablation, respectively. Similarly, in the ``Pred.'' column, $b/a$ denotes that out of $b$ original predictions, $a$ remained unchanged after the target expressions were ablated.}
\end{table}

%% file: tables/ab_context.tex
\begin{table}[t]
\centering
\small
\begin{tabular}{@{}llll@{}}
\toprule
Model     & \multicolumn{2}{c}{Llama-3-base} & Swallow-base \\ \cmidrule(l){2-3} \cmidrule(l){4-4}
Language & EN & JA & JA  \\ \cmidrule(l){2-2} \cmidrule(l){3-3} \cmidrule(l){4-4}
         & Before/After  & Before/After & Before/After     \\ \midrule
Ackn.    & 0.86/0.71    & 0.86/0.70          & 0.86/0.74               \\
Alte.    & 0.88/0.65    & 0.79/0.67          & 0.90/0.83               \\
Clari-Q. & 0.70/0.33   & 0.68/0.21           & 0.68/0.52               \\
Comm.    & 0.59/0.40   & 0.58/0.46           & 0.59/0.48               \\
Cond.    & 0.00/0.00   & 0.00/0.00           & 0.00/0.00               \\
Conf-Q.  & 0.93/0.82   & 0.92/0.90           & 0.91/0.90               \\
Contin.  & 0.51/0.27   & 0.48/0.21           & 0.46/0.24               \\
Contr.   & 0.77/0.70   & 0.73/0.66           & 0.77/0.70               \\
Corr.    & 0.79/0.33   & 0.79/0.55           & 0.81/0.57               \\
Elab.    & 0.77/0.67   & 0.77/0.59           & 0.76/0.71               \\
Expl.    & 0.00/0.00   & 0.00/0.00           & 0.00/0.00               \\
Narr.    & 0.90/0.03   & 0.89/0.04           & 0.89/0.02               \\
QAP.     & 0.82/0.16   & 0.83/0.29           & 0.85/0.30               \\
QElab.   & 0.52/0.42   & 0.34/0.27           & 0.42/0.33               \\
Resu.    & 0.91/0.26   & 0.91/0.53           & 0.91/0.59               \\
Sequ.    & 0.25/0.22   & 0.29/0.10           & 0.00/0.16               \\ \midrule
Weighted & 0.81/0.38   & 0.81/0.47           & 0.81/0.52               \\ \bottomrule
\end{tabular}
\caption{\label{ab_context}F1-scores before and after ablation of input discourse relations. The ``Weighted'' row indicates the overall performance weighted by relation frequency. }
\end{table}

%% file: tables/ab_mask.tex
\begin{table}[t]
\centering
\small
\begin{tabular}{@{}llll@{}}
\toprule
Model                                          & \multicolumn{2}{c}{Llama-3-base} & Swallow-base \\ \cmidrule(l){2-3} \cmidrule(l){4-4}
Language & EN & JA & JA \\ \cmidrule(l){2-2} \cmidrule(l){3-3} \cmidrule(l){4-4}
& Before/After & Before/After &Before/After \\ \midrule
Ackn.    & 0.86/0.81  & 0.86/0.80  & 0.86/0.70\\
Alte.    & 0.88/0.88  & 0.79/0.82  & 0.90/0.91           \\
Clari-Q. & 0.70/0.62  & 0.68/0.65  & 0.68/0.66            \\
Comm.    & 0.59/0.54  & 0.58/0.52  & 0.59/0.54           \\
Cond.    & 0.00/0.00  & 0.00/0.00  & 0.00/0.00           \\
Conf-Q.  & 0.93/0.88  & 0.92/0.89  & 0.91/0.89          \\
Contin.  & 0.51/0.40  & 0.48/0.44  & 0.46/0.40    \\
Contr.   & 0.77/0.77  & 0.73/0.59  & 0.77/0.73     \\
Corr.    & 0.79/0.70  & 0.79/0.57  & 0.81/0.68      \\
Elab.    & 0.77/0.72  & 0.77/0.75  & 0.76/0.71           \\
Expl.    & 0.00/0.00  & 0.00/0.00  & 0.00/0.11           \\
Narr.    & 0.90/0.73  & 0.89/0.78  & 0.89/0.73           \\
QAP.     & 0.82/0.79  & 0.83/0.71  & 0.85/0.73           \\
QElab.   & 0.52/0.36  & 0.34/0.32  & 0.42/0.36           \\
Resu.    & 0.91/0.86  & 0.91/0.87  & 0.91/0.84           \\
Sequ.    & 0.25/0.15  & 0.29/0.09  & 0.00/0.00           \\ \midrule
Weighted & 0.81/0.75  & 0.81/0.75  & 0.81/0.74           \\ \bottomrule
\end{tabular}
\caption{\label{ab_mask_context} F1-scores before and after masking of the content of the context clauses. The ``Weighted'' indicates the overall performance weighted by relation frequency.}
\end{table}

%% file: tables/translation_ex.tex
\begin{table}[h]
\centering
\small
\begin{tabular}{@{}clp{4.5cm}@{}}
\toprule
 Index & Speaker   & Text                                      \\ \midrule
 \multirow{2}{*}{2}     & \multirow{2}{*}{Builder}   & ``lets do this''   \\ 
 &  & ``やろうか'' \\ \midrule
 \multirow{2}{*}{3}     &  \multirow{2}{*}{Architect} & ``place a red block in front of the blue block'' \\
 &  & ``青いブロックの前に赤いブロックを置いてください'' \\ \midrule
\multirow{2}{*}{4}     & \multirow{2}{*}{Architect} & ``along the line toward the purple block''       \\
&   & ``紫のブロックに向かう線に沿って'' \\ \midrule
 \multirow{2}{*}{5}     & \multirow{2}{*}{Builder}   & ``place red 4 1 0''                              \\
 &  &   ``設置 赤 4 1 0''\\ \bottomrule
\end{tabular}
\caption{\label{translations}Examples of translation of MSDC. The sentence above is the original English, while the sentence below is the Japanese translation.}
\end{table}

%% file: tables/error_ex.tex
\begin{table}[ht]
\centering
\small
\begin{tabular}{@{}lp{6cm}@{}}
\toprule
 Speaker & Text   \\ \midrule
 Builder & ``any color''  \\
 Builder & ``or still orange?'' \\ \midrule
 Builder & ``どんな色でもいいよ'' \\
 Builder & ``それともオレンジのままでいい?'' \\ \midrule \midrule
 Architect & ``on the top orange block'' \\
 Architect & ``put a yellow block to the left of it'' \\ \midrule
 Architect &``一番上のオレンジのブロックの上に''\\
 Architect &``その左側に黄色のブロックを置いてください''\\ \midrule \midrule
 Architect & ``On the top and bottom'' \\
 Architect & ``we are going to add 2 block extensions'' \\ \midrule
 Architect &``上と下に，''\\
Architect & ``上と下に，ブロックを二つ分突き出させます''\\ \bottomrule
\end{tabular}
\caption{\label{error}Examples of translation errors. In each example, the upper segment provides the original English excerpt, and the lower segment presents its corresponding Japanese translation.}
\end{table}

%% file: tables/hyperparameters.tex
\begin{table}[h]
\centering
\begin{tabular}{@{}ll@{}}
\toprule
\multicolumn{2}{c}{GPUs}             \\ \midrule
\multicolumn{2}{c}{NVIDIA Hopper H100} \\ \midrule\midrule
\multicolumn{2}{c}{Hyperparameters} \\ \midrule
training epochs             & 3     \\
batch size                  & 1     \\
optimizer                   & Adam     \\
learning rate               & 2e-4   \\
\multirow{2}{*}{learning rate scheduler}     & linear warm-up and     \\
                                            & cosine annealing \\
warm-up ratio               & 0.03    \\
gradient clipping           &  0.3    \\
LoRA r                      & 64     \\
LoRA (alpha)                & 16     \\
LoRA dropout ratio          & 0.1     \\
\multirow{2}{*}{LoRA target modules}         &  Only Attention Blocks\\ 
                                        & (q\_proj, v\_proj)    \\
quantization      & 4-bit NormalFloat \\ \bottomrule
\end{tabular}
\caption{\label{tab:model-details}Hyperparameters for the training.}
\end{table}

%% file: tables/results_all_MSDC.tex
\begin{table*}[h]
\centering
\begin{tabular}{@{}lllllllllll@{}}
\toprule
                      & \multicolumn{3}{c}{EN (Llama-3-base)}                          & \multicolumn{3}{c}{JA (Llama-3-base)}                         & \multicolumn{3}{c}{JA (Swallow-base)}                 &                        \\ \cmidrule(l){2-4} \cmidrule(l){5-7} \cmidrule(l){8-10}
 Relation                     & Prec.            & Recall             & F1            & Prec.            & Recall              & F1             & Prec.            & Recall               & F1             & Supports            \\ \midrule
Ackn.       & 0.86                 & 0.85                 & \bf{0.86}                 & 0.86                 & 0.87                 & \bf{0.86}                 & 0.85                 & 0.85                 & 0.85                 & 792                  \\
Alte.         & 0.85                 & 0.96                 & \bf{0.90}                 & 0.70                & 0.96                 & 0.81                 & 0.85                 & 0.96                 & \bf{0.90}                 & 24                    \\
Clari-Q.       & 0.67                 & 0.72                 & \bf{0.69}                 & 0.66                 & 0.69                 & 0.67                 & 0.66                 & 0.70                 & 0.68                 & 134                  \\
Comm.               & 0.62                 & 0.54                 & 0.58                 & 0.63                 & 0.54                 & 0.58                 & 0.65                 & 0.53                 & 0.58                 & 382                  \\
Cond.          & 0.00                 & 0.00                 & 0.00                 & 0.00                 & 0.00                 & 0.00                 & 0.00                 & 0.00                 & 0.00                 & 13                   \\
Conf-Q.        & 0.97                 & 0.89                 & \bf{0.93}                 & 0.93                 & 0.88                 & 0.91                 & 0.93                 & 0.88                 & 0.90                 & 159                  \\
Contin.          & 0.40                 & 0.52                 & \bf{0.45}                 & 0.41                 & 0.51                 & \bf{0.45}                 & 0.39                 & 0.45                 & 0.42                 & 321                   \\
Contr.             & 0.90                 & 0.66                 & 0.76                 & 0.87                 & 0.68                 & 0.76                 & 0.91                 & 0.74                 & \bf{0.81}                 & 68                    \\
Corr.            & 0.73                 & 0.71                 & 0.72                 & 0.77                 & 0.72                 & \bf{0.75}                 & 0.81                 & 0.69                 & 0.74                 & 403                   \\
Elab.          & 0.78                 & 0.76                 & \bf{0.77}                 & 0.77                 & 0.76                 & 0.76                 & 0.73                 & 0.78                 & 0.75                 & 720                   \\
Expl.        & 0.00                 & 0.00                 & 0.00                 & 0.00                 & 0.00                 & 0.00                 & 0.00                 & 0.00                 & 0.00                 & 16                    \\
Narr.             & 0.82                 & 0.82                 & \bf{0.82}                 & 0.80                 & 0.81                 & 0.80                 & 0.81                 & 0.81                 & 0.81                 & 693                   \\
QAP. & 0.85                 & 0.78                 & 0.81                 & 0.83                 & 0.80                 & 0.81                 & 0.83                 & 0.81                 & \bf{0.82}                 & 290                    \\
QElab.         & 0.48                 & 0.39                 & \bf{0.43}                 & 0.41                 & 0.37                 & 0.39                 & 0.44                 & 0.39                 & 0.42                 & 38                   \\
Resu.                & 0.90                 & 0.91                 & 0.90                 & 0.90                 & 0.90                 & 0.90                 & 0.90                 & 0.90                 & 0.90                 & 1765                \\
Sequ.              & 0.50                 & 0.17                 & 0.25                 & 1.00                 & 0.17                 & \bf{0.29}                 & 0.00                 & 0.00                 & 0.00                 & 6                      \\ \midrule \midrule
Weighted Average      & 0.80 & 0.79 & 0.79    & 0.80 & 0.79 & 0.79  & 0.79    & 0.78 & 0.79 \\ \bottomrule
\end{tabular}
\caption{\label{result_all_msdc}Performance of the constructed SDRT parsers. The ``Supports'' column indicates the number of instances for each relation in the test data. The Relation labels are shown in abbreviated form. The best F1-scores among the models are highlighted in bold.}
\end{table*}

%% file: tables/results_all_STAC.tex
\begin{table*}[h]
\centering
\begin{tabular}{@{}lllllllllll@{}}
\toprule
                      & \multicolumn{3}{c}{EN (Llama-3-base)}                          & \multicolumn{3}{c}{JA (Llama-3-base)}                         & \multicolumn{3}{c}{JA (Swallow-base)}                 &                        \\ \cmidrule(l){2-4} \cmidrule(l){5-7} \cmidrule(l){8-10}
 Relation   & Prec.  & Recall  & F1   & Prec.  & Recall   & F1                 & Prec.  & Recall  & F1   & Supports             \\ \midrule
Ackn.       & 0.64   & 0.58    & 0.61  & 0.63   & 0.61     & 0.62               & 0.65   & 0.66   & \bf{0.65}  & 173                   \\
Alte.       & 0.75   & 0.60    & \bf{0.67}  & 0.71   & 0.50     & 0.59               & 0.83   & 0.50    & 0.62  & 10                    \\
Back.       & 0.00   & 0.00    & 0.00  & 1.00   & 0.11     & \bf{0.20}                & 0.00   & 0.00    & 0.00  & 9 \\
Clari-Q.    & 0.44   & 0.58    & \bf{0.50}  & 0.39   & 0.54     & 0.46               & 0.46   & 0.54    & \bf{0.50}  & 48              \\
Comm.       & 0.58   & 0.62    & 0.60  & 0.58   & 0.63     & 0.60               & 0.58   & 0.62    & 0.60  & 222                  \\
Cond.       & 0.71   & 0.42    & 0.53  & 0.82   & 0.75     & \bf{0.78}               & 0.86   & 0.50    & 0.63  & 12                   \\
Contin.     & 0.27   & 0.30    & 0.29  & 0.23   & 0.28     & 0.25               & 0.34   & 0.38    & \bf{0.36}  & 69                   \\
Contr.      & 0.53   & 0.47    & \bf{0.50}  & 0.47   & 0.40     & 0.43               & 0.52   & 0.47    & 0.49  & 60                    \\
Corr.       & 0.50   & 0.30    & \bf{0.38}  & 0.36   & 0.17     & 0.24               & 0.38   & 0.26    & 0.31  & 23                   \\
Elab.       & 0.50   & 0.51    & \bf{0.51}  & 0.48   & 0.44     & 0.46               & 0.43   & 0.45    & 0.44  & 100                   \\
Expl.       & 0.40   & 0.38    & \bf{0.39}  & 0.36   & 0.38     & 0.37               & 0.39   & 0.36    & 0.38  & 55                    \\
Narr.       & 0.00   & 0.00    & 0.00  & 0.00   & 0.00     & 0.00               & 0.00   & 0.00    & 0.00  & 8                   \\
Para.       & 0.50   & 0.26    & \bf{0.34}  & 0.45   & 0.22     & 0.29               & 0.50   & 0.26    & \bf{0.34}  & 23 \\
QAP.        & 0.80   & 0.83    & \bf{0.81}  & 0.82   & 0.81     & \bf{0.81}               & 0.77   & 0.79    & 0.78  & 273\\
QElab.      & 0.58   & 0.42    & 0.49  & 0.58   & 0.49     & \bf{0.53}               & 0.54   & 0.46    & 0.50  & 84                   \\
Resu.       & 0.67   & 0.64    & \bf{0.65}  & 0.63   & 0.60     & 0.61               & 0.65   & 0.58    & 0.61  & 145                \\
Sequ.       & 0.82   & 0.75    & \bf{0.78}  & 0.82   & 0.75     & \bf{0.78}               & 0.78   & 0.73    & 0.75  & 124                      \\ \midrule \midrule
Weighted Average   & 0.62 & 0.60 & \bf{0.60}  & 0.61 & 0.59 & 0.59  & 0.60    & 0.59 & 0.59 & \\ \bottomrule
\end{tabular}
\caption{\label{result_all_STAC}Performance of the constructed SDRT parsers on STAC. The ``Supports'' column indicates the number of instances for each relation in the test data. The Relation labels are shown in abbreviated form. The best F1-scores among the models are highlighted in bold.}
\end{table*}

%% file: tables/eval_summary_subset_llamipa.tex
\begin{table*}[t]
\centering
\resizebox{\textwidth}{!}{%
\begin{tabular}{@{}lccccccccccccccccccc@{}}
\toprule
         & \multicolumn{2}{c}{Llama-3} & Swallow-3 & \multicolumn{6}{c}{Llama-3.1}                                                      & \multicolumn{3}{c}{Swallow-3.1} & \multicolumn{6}{c}{GPT-5.5}                                                        \\ \cmidrule(lr){2-3} \cmidrule(lr){4-4} \cmidrule(lr){5-10} \cmidrule(lr){11-13}\cmidrule(lr){14-19}
         & \multicolumn{3}{c}{Llamipa}           & \multicolumn{2}{c}{0shot} & \multicolumn{2}{c}{1shot} & \multicolumn{2}{c}{4shot} & 0shot    & 1shot    & 4shot    & \multicolumn{2}{c}{0shot} & \multicolumn{2}{c}{1shot} & \multicolumn{2}{c}{4shot} \\ \cmidrule(l){2-4}  \cmidrule(lr){5-6}\cmidrule(lr){7-8}\cmidrule(lr){9-10}\cmidrule(lr){11-11}\cmidrule(lr){12-12}\cmidrule(lr){13-13}\cmidrule(lr){14-15}\cmidrule(lr){16-17}\cmidrule(lr){18-19}
Relation & EN           & JA          & JA       & EN          & JA          & EN          & JA          & EN          & JA          & JA       & JA       & JA       & EN          & JA          & EN          & JA          & EN          & JA  & Support        \\ \midrule
Ackn.    & 0.86         & 0.86        & 0.85     & 0.56        & 0.56        & 0.47        & 0.43        & 0.40        & 0.43        & 0.51     & 0.36     & 0.15     & 0.55        & 0.56        & 0.76        & 0.76        & 0.78        & 0.78     & 792   \\
Alte.    & 0.90         & 0.81        & 0.90     & 0.00        & 0.00        & 0.00        & 0.00        & 0.32        & 0.42        & 0.00     & 0.00     & 0.13     & 0.68        & 0.76        & 0.88        & 0.65        & 0.88        & 0.88     & 24   \\
Clari-Q. & 0.69         & 0.67        & 0.68     & 0.06        & 0.08        & 0.22        & 0.24        & 0.29        & 0.27        & 0.01     & 0.19     & 0.08     & 0.44        & 0.40        & 0.60        & 0.58        & 0.56        & 0.60   & 134     \\
Comm.    & 0.58         & 0.58        & 0.58     & 0.16        & 0.16        & 0.28        & 0.23        & 0.25        & 0.12        & 0.19     & 0.14     & 0.03     & 0.33        & 0.34        & 0.45        & 0.47        & 0.52        & 0.49    & 382    \\
Cond.    & 0.00         & 0.00        & 0.00     & 0.00        & 0.00        & 0.00        & 0.00        & 0.00        & 0.00        & 0.00     & 0.00     & 0.00     & 0.20        & 0.22        & 0.19        & 0.22        & 0.29        & 0.00    & 13     \\
Conf-Q.   & 0.93         & 0.91        & 0.90     & 0.04        & 0.07        & 0.58        & 0.56        & 0.49        & 0.56        & 0.01     & 0.55     & 0.14     & 0.76        & 0.78        & 0.88        & 0.89        & 0.88        & 0.90    & 159    \\
Contin.  & 0.45         & 0.45        & 0.42     & 0.07        & 0.04        & 0.06        & 0.06        & 0.11        & 0.07        & 0.06     & 0.04     & 0.03     & 0.15        & 0.11        & 0.24        & 0.26        & 0.41        & 0.40    & 321    \\
Contr.   & 0.76         & 0.76        & 0.81     & 0.07        & 0.03        & 0.11        & 0.10        & 0.00        & 0.07        & 0.04     & 0.10     & 0.00     & 0.52        & 0.55        & 0.69        & 0.77        & 0.78        & 0.73   & 68     \\
Corr.    & 0.72         & 0.75        & 0.74     & 0.02        & 0.03        & 0.09        & 0.08        & 0.08        & 0.07        & 0.03     & 0.08     & 0.01     & 0.26        & 0.22        & 0.47        & 0.49        & 0.60        & 0.60    & 403    \\
Elab.    & 0.77         & 0.76        & 0.75     & 0.35        & 0.32        & 0.40        & 0.37        & 0.39        & 0.41        & 0.30     & 0.38     & 0.12     & 0.59        & 0.55        & 0.70        & 0.72        & 0.74        & 0.73     & 720   \\
Expl.    & 0.00         & 0.00        & 0.00     & 0.01        & 0.02        & 0.00        & 0.04        & 0.02        & 0.05        & 0.01     & 0.04     & 0.00     & 0.17        & 0.10        & 0.20        & 0.15        & 0.19        & 0.16    & 16    \\
Narr.    & 0.82         & 0.80        & 0.81     & 0.00        & 0.00        & 0.00        & 0.00        & 0.02        & 0.01        & 0.00     & 0.00     & 0.00     & 0.00        & 0.00        & 0.11        & 0.17        & 0.68        & 0.68    & 693    \\
QAP.     & 0.81         & 0.81        & 0.82     & 0.05        & 0.03        & 0.43        & 0.41        & 0.42        & 0.46        & 0.03     & 0.37     & 0.11     & 0.65        & 0.63        & 0.79        & 0.78        & 0.81        & 0.79    &290    \\
QElab.   & 0.43         & 0.39        & 0.42     & 0.05        & 0.00        & 0.00        & 0.00        & 0.00        & 0.00        & 0.00     & 0.00     & 0.00     & 0.11        & 0.10        & 0.05        & 0.12        & 0.16        & 0.17    & 38    \\
Resu.    & 0.90         & 0.90        & 0.90     & 0.38        & 0.35        & 0.54        & 0.50        & 0.50        & 0.52        & 0.30     & 0.51     & 0.12     & 0.49        & 0.51        & 0.67        & 0.73        & 0.85        & 0.85    & 1765   \\
Sequ.    & 0.25         & 0.29        & 0.00     & 0.00        & 0.00        & 0.00        & 0.00        & 0.00        & 0.00        & 0.00     & 0.00     & 0.00     & 0.00        & 0.00        & 0.00        & 0.00        & 0.00        & 0.00     & 6   \\ \midrule
\makecell{Weighted \\ Average} & 0.79         & 0.79        & 0.79     & 0.26        & 0.24        & 0.35        & 0.32        & 0.33        & 0.33        & 0.22     & 0.30     & 0.09     & 0.42        & 0.41        & 0.57        & 0.60        & 0.73        & 0.72        \\ \bottomrule
\end{tabular}
}
\caption{\label{eval_sum_subset}F1-scores for Llamipa and prompt-based methods, evaluated on a subset of gold relations from the MSDC corpus (cf. Appendix~\ref{App: Evaluation on All Relation of MSDC}). "Support" represents the number of instances per relation in the test data, and discourse relations are abbreviated.}

\end{table*}

%% file: tables/eval_all_rel.tex
\begin{table*}[t]
\resizebox{\textwidth}{!}{%
\begin{tabular}{@{}lccccccccccccccccccc@{}}
\toprule
         & \multicolumn{2}{c}{Llama-3} & Swallow-3 & \multicolumn{6}{c}{Llama-3.1}                                                      & \multicolumn{3}{c}{Swallow-3.1} & \multicolumn{6}{c}{GPT-5.5}                                                        \\ \cmidrule(lr){2-3} \cmidrule(lr){4-4} \cmidrule(lr){5-10} \cmidrule(lr){11-13}\cmidrule(lr){14-19}
         & \multicolumn{3}{c}{Llamipa}           & \multicolumn{2}{c}{0shot} & \multicolumn{2}{c}{1shot} & \multicolumn{2}{c}{4shot} & 0shot    & 1shot    & 4shot    & \multicolumn{2}{c}{0shot} & \multicolumn{2}{c}{1shot} & \multicolumn{2}{c}{4shot} \\ \cmidrule(l){2-4}  \cmidrule(lr){5-6}\cmidrule(lr){7-8}\cmidrule(lr){9-10}\cmidrule(lr){11-11}\cmidrule(lr){12-12}\cmidrule(lr){13-13}\cmidrule(lr){14-15}\cmidrule(lr){16-17}\cmidrule(lr){18-19}
Relation & EN           & JA          & JA       & EN          & JA          & EN          & JA          & EN          & JA          & JA       & JA       & JA       & EN          & JA          & EN          & JA          & EN          & JA  & Support        \\ \midrule
Ackn.    & 0.86         & 0.86        & 0.85     & 0.56        & 0.56        & 0.47        & 0.43        & 0.40        & 0.43        & 0.51     & 0.36     & 0.15     & 0.55        & 0.56        & 0.76        & 0.76        & 0.78        & 0.78     & 792   \\
Alte.    & 0.90         & 0.81        & 0.90     & 0.00        & 0.00        & 0.00        & 0.00        & 0.32        & 0.42        & 0.00     & 0.00     & 0.13     & 0.68        & 0.76        & 0.88        & 0.65        & 0.88        & 0.88    & 24    \\
Clari-Q. & 0.69         & 0.67        & 0.68     & 0.06        & 0.08        & 0.22        & 0.24        & 0.29        & 0.27        & 0.01     & 0.19     & 0.08     & 0.44        & 0.40        & 0.60        & 0.58        & 0.56        & 0.60    & 134    \\
Comm.    & 0.54         & 0.54        & 0.54     & 0.15        & 0.14        & 0.26        & 0.21        & 0.23        & 0.11        & 0.18     & 0.13     & 0.03     & 0.31        & 0.32        & 0.42        & 0.43        & 0.48        & 0.46    &434    \\
Cond.    & 0.00         & 0.00        & 0.00     & 0.00        & 0.00        & 0.00        & 0.00        & 0.00        & 0.00        & 0.00     & 0.00     & 0.00     & 0.17        & 0.27        & 0.16        & 0.18        & 0.24        & 0.10    & 17    \\
Conf-Q.   & 0.93         & 0.91        & 0.90     & 0.04        & 0.07        & 0.58        & 0.56        & 0.49        & 0.56        & 0.01     & 0.55     & 0.14     & 0.76        & 0.78        & 0.88        & 0.89        & 0.88        & 0.90     & 159   \\
Contin.  & 0.45         & 0.45        & 0.42     & 0.07        & 0.04        & 0.06        & 0.06        & 0.11        & 0.07        & 0.06     & 0.04     & 0.03     & 0.15        & 0.11        & 0.24        & 0.26        & 0.41        & 0.40    & 321    \\
Contr.   & 0.76         & 0.76        & 0.81     & 0.07        & 0.03        & 0.11        & 0.10        & 0.00        & 0.07        & 0.04     & 0.10     & 0.00     & 0.52        & 0.55        & 0.69        & 0.77        & 0.78        & 0.73   & 68     \\
Corr.    & 0.72         & 0.75        & 0.74     & 0.02        & 0.03        & 0.09        & 0.08        & 0.08        & 0.07        & 0.03     & 0.08     & 0.01     & 0.26        & 0.22        & 0.47        & 0.49        & 0.60        & 0.60  & 405      \\
Elab.    & 0.77         & 0.76        & 0.75     & 0.35        & 0.32        & 0.40        & 0.37        & 0.39        & 0.41        & 0.30     & 0.38     & 0.12     & 0.59        & 0.55        & 0.70        & 0.72        & 0.74        & 0.73     & 720   \\
Expl.    & 0.00         & 0.00        & 0.00     & 0.01        & 0.02        & 0.00        & 0.04        & 0.02        & 0.05        & 0.01     & 0.04     & 0.00     & 0.17        & 0.10        & 0.20        & 0.15        & 0.19        & 0.16    & 16    \\
Narr.    & 0.79         & 0.78        & 0.78     & 0.00        & 0.00        & 0.00        & 0.00        & 0.02        & 0.01        & 0.00     & 0.00     & 0.00     & 0.00        & 0.00        & 0.11        & 0.17        & 0.68        & 0.68   & 739     \\
QAP.     & 0.81         & 0.81        & 0.82     & 0.05        & 0.03        & 0.43        & 0.41        & 0.42        & 0.46        & 0.03     & 0.37     & 0.11     & 0.65        & 0.63        & 0.79        & 0.78        & 0.81        & 0.79   & 290     \\
QElab.   & 0.43         & 0.39        & 0.42     & 0.05        & 0.00        & 0.00        & 0.00        & 0.00        & 0.00        & 0.00     & 0.00     & 0.00     & 0.11        & 0.10        & 0.05        & 0.12        & 0.16        & 0.17   & 38     \\
Resu.    & 0.90         & 0.90        & 0.90     & 0.38        & 0.35        & 0.54        & 0.50        & 0.50        & 0.52        & 0.30     & 0.51     & 0.12     & 0.49        & 0.51        & 0.67        & 0.73        & 0.85        & 0.85   & 1765     \\
Sequ.    & 0.25         & 0.29        & 0.00     & 0.00        & 0.00        & 0.00        & 0.00        & 0.00        & 0.00        & 0.00     & 0.00     & 0.00     & 0.00        & 0.00        & 0.00        & 0.00        & 0.00        & 0.00   & 6     \\ \midrule
\makecell{Weighted \\ Average} & 0.78         & 0.78        & 0.78     & 0.25        & 0.24        & 0.34        & 0.32        & 0.32        & 0.33        & 0.21     & 0.30     & 0.09     & 0.41        & 0.41        & 0.57        & 0.60        & 0.72        & 0.72        \\ \bottomrule
\end{tabular}
}
\caption{\label{eval_all_rel}F1-scores for Llamipa and prompt-based methods on the MSDC corpus, evaluated across all gold relations.
"Support" represents the number of instances per relation in the test data, and discourse relations are abbreviated.}
\end{table*}